\documentclass[runningheads]{llncs}

\usepackage[mobile]{eccv}

\usepackage{eccvabbrv}
\usepackage{graphicx}%
\usepackage{multirow}%
\usepackage{amsmath,amssymb,amsfonts}%
\usepackage{mathrsfs}%
\usepackage[title]{appendix}%
\usepackage{xcolor}%
\usepackage{textcomp}%
\usepackage{manyfoot}%
\usepackage{booktabs}%
\usepackage{algorithm}%
\usepackage{algorithmicx}%
\usepackage{algpseudocode}%
\usepackage{listings}%
\usepackage{wrapfig}

\usepackage{graphicx}
\usepackage{amsmath}
\usepackage{amssymb}
\usepackage{booktabs}
\usepackage{amsfonts,amssymb}
\usepackage[accsupp]{axessibility}
\usepackage{multirow}
\usepackage{float}
\usepackage{caption}
\usepackage{subcaption}
\usepackage{fontawesome}
\usepackage{tabularx,caption}
\usepackage[rightcaption]{sidecap}
\usepackage{xcolor}
\usepackage{colortbl}
\usepackage{psfrag}
\usepackage[percent]{overpic}
\usepackage{url}
\usepackage[accsupp]{axessibility}
\usepackage{tablefootnote}

\definecolor{mygray}{gray}{.9}
\makeatletter
\renewcommand\subsubsection{\@startsection{subsubsection}{3}{0pt}%
  {-18pt \@plus -4pt \@minus -4pt}%
  {0.5em \@plus 0.22em \@minus 0.1em}%
  {\normalfont\normalsize\bfseries}}

\makeatother

\usepackage{graphicx}
\usepackage{booktabs}

\usepackage[accsupp]{axessibility}  

\usepackage[pagebackref,breaklinks,colorlinks,citecolor=eccvblue]{hyperref}

\usepackage{orcidlink}

\begin{document}

\title{Deep Makeup Guided Facial Privacy Prior} 
\title{Makeup-Guided Facial Privacy Protection via Untrained Neural Network Priors}

\titlerunning{Makeup-Guided Facial Privacy Protection}

\author{Fahad Shamshad\inst{} \and
Muzammal Naseer\inst{} \and
Karthik Nandakumar\inst{}}

\authorrunning{~Shamshad et al.}

\institute{Mohamed bin Zayed University of Artificial Intelligence, UAE} 

\maketitle

\begin{abstract}
Deep learning-based face recognition (FR) systems pose significant privacy risks by tracking users without their consent.  While adversarial attacks can protect privacy, they often produce visible artifacts compromising user experience. To mitigate this issue, recent facial privacy protection approaches advocate embedding adversarial noise into the natural looking makeup styles. However, these methods require training on large-scale makeup datasets that are not always readily available. In addition, these approaches also suffer from dataset bias. For instance, training on makeup data that predominantly contains female faces could compromise protection efficacy for male faces. To handle these issues, we propose a test-time optimization approach that solely optimizes an untrained neural network to transfer makeup style from a reference to a source image in an adversarial manner. We introduce two key modules: a correspondence module that aligns regions between reference and source images in latent space, and a decoder with conditional makeup layers. The untrained decoder, optimized via carefully designed structural and makeup consistency losses, generates a protected image that resembles the source but incorporates adversarial makeup to deceive FR models. As our approach does not rely on training with makeup face datasets, it avoids potential male/female dataset biases while providing effective protection. We further extend the proposed approach to videos by leveraging on temporal correlations.  Experiments on benchmark datasets demonstrate superior performance in face verification and identification tasks and effectiveness against commercial FR systems. Our code and models will be available at \href{https://github.com/fahadshamshad/deep-facial-privacy-prior}{\color{Magenta}{{https://github.com/fahadshamshad/deep-facial-privacy-prior}}}.

  \keywords{Facial privacy protection \and adversarial makeup transfer \and face recognition \and black-box attacks}
\end{abstract}

\section{Introduction} \label{sec:intro}

Face recognition (FR) systems are widely used in various applications, including biometrics~\cite{meden2021privacy}, security~\cite{wang2017face}, and criminal investigations~\cite{rezende2020facial}. The advent of deep learning has significantly improved the performance of FR systems~\cite{parkhi2015deep,wang2021deep}. While deep learning-based FR systems offer many potential benefits, they also pose significant privacy risks, enabling potential mass surveillance by governments and private organizations, particularly on social media platforms~\cite{ahern2007over,wenger2021sok,kamgar2011toward}. The prevalence of proprietary, black-box FR models further emphasizes the urgent need for effective facial privacy protection.

\begin{table*}[t]
\begin{center}
\caption{\footnotesize Comparison of facial privacy protection methods across output naturalness, black-box transferability, verification/identification performance, and unrestricted (semantically meaningful) examples, and use of reference images.}
\small 
\label{table:comparison}
\vspace{-2mm}
\setlength{\tabcolsep}{2.0pt}
\scalebox{0.75}{
\begin{tabular}{l | c | c | c | c | c }
\toprule[0.15em]
\rowcolor{mygray} &Adv-Makeup~\cite{DBLP:conf/ijcai/YinWYGKDLL21} &TIP-IM~\cite{yang2021towards}& AMT-GAN~\cite{hu2022protecting} & CLIP2Protect~\cite{shamshad2023clip2protect}& \texttt{DFPP} (Ours)\\  
\midrule[0.15em]
Natural outputs  &  Yes & Partially  & Partially &Yes & \textbf{Yes}  \\
Black box  &  Yes & Yes  & Yes & Yes & \textbf{Yes}  \\
Verification  &  Yes & No  & Yes & Yes & \textbf{Yes}  \\
Identification  &  No & Yes  & No & Yes & \textbf{Yes}\\
Unrestricted  &  Yes & No  & Yes & Yes & \textbf{Yes}\\
Reference Img.  &  Yes & No  & Yes & No & \textbf{Yes}\\
\bottomrule[0.1em]
\end{tabular}}
\end{center}\vspace{-1.5em}
\end{table*}

\begin{SCfigure}[][t]
\centering
\includegraphics[clip, trim=0cm 0.0cm 0cm 0cm,width=0.55\textwidth]{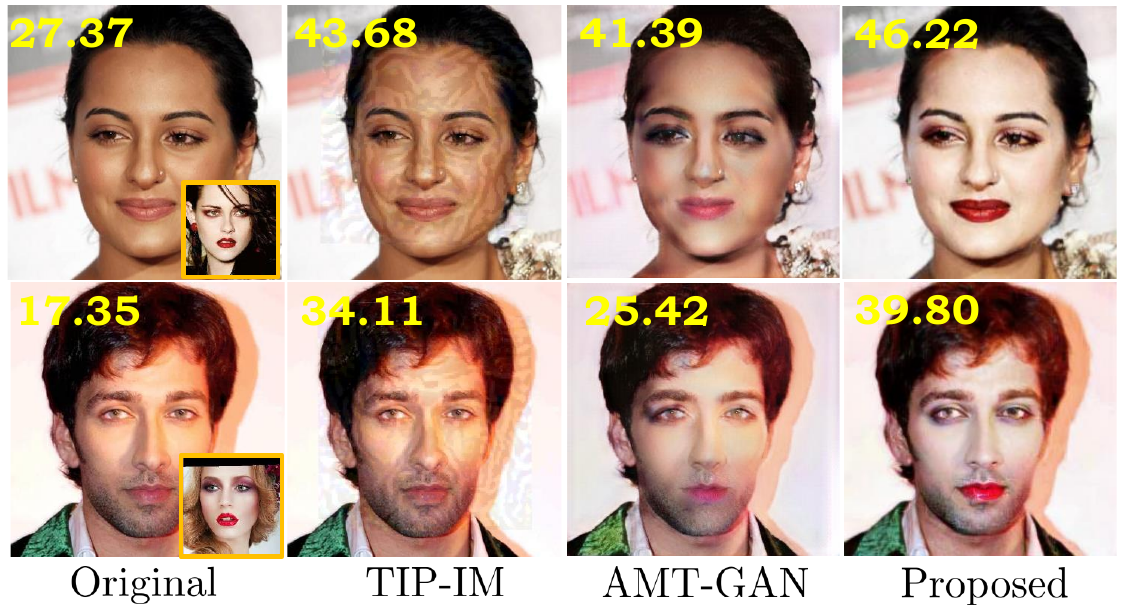}
\vspace{0em}
\caption{\footnotesize Our approach generates more natural protected faces for deceiving black-box face recognition systems, outperforming TIP-IM~\cite{yang2021towards} and AMT-GAN~\cite{hu2022protecting}. The yellow text at the top-left shows the confidence score (higher is better) from a commercial API when matching the protected image to a false target identity.} 
\label{fig:title_fig}
\end{SCfigure}

Recent studies have explored adversarial attacks to protect facial privacy, but these often struggle to balance naturalness and privacy protection~\cite{yang2021towards}. Noise-based techniques typically produce visible artifacts~\cite{xiao2021improving,yang2020design}, as shown in Fig.~\ref{fig:title_fig}.
Unrestricted adversarial examples offer better perceptual realism while maintaining privacy~\cite{bhattad2019unrestricted,zhao2020towards,song2018constructing,liu2022towards,yuan2022natural}, with makeup-based methods embedding adversarial perturbations naturally through makeup effects~\cite{DBLP:conf/ijcai/YinWYGKDLL21,hu2022protecting}.  These methods use generative adversarial networks trained on large makeup datasets to transfer makeup from a reference image to a user's face, imitating a target identity. 
Importantly, these makeup-based approaches learn effective \textit{image priors} by capturing natural image statistics from large-scale makeup datasets. Nevertheless, despite their effectiveness, existing adversarial makeup transfer methods suffer from several limitations, as discussed next.

\textit{\textbf{First}}, training on large makeup datasets is required to capture makeup statistics. These datasets are not only difficult to acquire, but also make these approaches susceptible to dataset bias, as prior information is generally limited to the statistics of the data used for training. 
\textit{\textbf{Second}}, recent test-time based makeup transfer approaches, such as CLIP2Protect, rely on pre-trained models like StyleGAN for generating protected images. This dependency makes these methods vulnerable to the inherent dataset biases of the pre-trained models, potentially leading to suboptimal performance across diverse demographics. Moreover, high-quality image generators are often hard to train.
\textit{\textbf{Third}}, adversarial toxicity can cause false matches in semantic correspondences, leading to unnatural makeup artifacts and changes in the perceived identity of the user image (see Fig.~\ref{fig:title_fig}). While some methods use textual makeup guidance, this can be limiting for complex styles, and users may prefer reference images for finer control.

To address these issues, we propose an encoder-decoder-based approach, Deep Facial Privacy Prior (\texttt{DFPP}), that solely optimizes the weights of a randomly initialized neural network at test-time for natural-looking adversarial makeup transfer. Our approach features a \textbf{robust correspondence module} for semantic alignment of reference and source images in the encoder's latent space, and a randomly initialized conditional decoder with \textbf{Adaptive Makeup Conditioning} (AMC) layers. We optimize the decoder parameters at test-time to generate protected samples that retain (\textbf{i}) the source's human-perceived identity, (\textbf{ii}) adopt the reference image's makeup style, and (\textbf{iii}) mimic the target image identity to evade black-box FR models. To achieve these stated objectives, we carefully designed a composite loss function with three key components: a \textbf{Structural Consistency Loss} that maintains source identity via patch-wise matching in a pre-trained ViT feature space; a \textbf{Makeup Loss} that facilitates effective makeup transfer by matching region-wise color distribution and global tone while preserving background regions; and an \textbf{Adversarial Loss} that ensures the protected sample's features match the target image in the FR model's feature space while distancing from the source image embedding.

Unlike recent methods, \texttt{DFPP} avoids the need for large-scale training on makeup datasets, effectively mitigating dataset bias. Extensive experiments in face verification and identification tasks, under both impersonation and dodging scenarios, show that \texttt{DFPP} effectively evades malicious black-box FR models and commercial APIs. Additionally, we demonstrate the effectiveness of \texttt{DFPP} in protecting videos. For videos, we leverage our test-time optimization by transferring weights learned from one frame to subsequent frames, achieving approximately 10 times computational efficiency without compromising privacy.

\section{Related Work} \label{sec:related_work}

\noindent \textbf{Adversarial Examples for Face Recognition Protection}:
Adversarial attacks have been widely used to protect users from unauthorized FR models. These approaches can be broadly categorized into noise-based and unrestricted adversarial examples. Noise-based methods include data poisoning~\cite{cherepanova2020lowkey,shan2020fawkes}, game theory~\cite{oh2017adversarial}, and privacy masks~\cite{zhong2022opom}, but often require multiple user images, access to training data, or are limited to white-box settings. Recent work like TIP-IM~\cite{yang2021towards} targets black-box models but produces perceptible noise.
Unrestricted Adversarial Examples (UAEs) aim to create less noticeable perturbations~\cite{bhattad2019unrestricted,zhao2020towards,song2018constructing,liu2022towards,yuan2022natural}. These include patch-based attacks creating wearable items including hats or colorful glasses~\cite{komkov2021advhat,sharif2019general,xiao2021improving}, but they often suffer from poor transferability and unnatural appearance. Generative model-based UAEs show promise but have limited performance in black-box settings~\cite{zhu2019generating,poursaeed2021robustness,kakizaki2019adversarial}.

\noindent \textbf{Makeup-based Facial Privacy Protection}: Recent approaches have leveraged makeup-based unrestricted attacks~\cite{zhu2019generating, yin2021adv,guetta2021dodging, pi2023adv,hu2022protecting} to deceive FR systems by embedding adversarial perturbations into natural makeup effects. However, these methods often require training on large makeup datasets, potentially introducing gender bias, and can produce undesirable artifacts when source and reference styles differ significantly. CLIP2Protect~\cite{shamshad2023clip2protect} introduced a two-stage, text-guided method to address some of these issues, but it still relied on pre-trained StyleGANs, making it susceptible to inherent dataset biases~\cite{karakas2022fairstyle,munoz2023uncovering}. Additionally, text-based prompts may not capture complex makeup styles as effectively as reference images. More recently, DiffAM~\cite{sun2024diffam} utilized pre-trained diffusion models for facial privacy protection in face verification scenarios, but still relied on a pre-trained generator. In contrast, our proposed approach \texttt{DFPP} eliminates dependency on pre-trained generative models, mitigating dataset bias issues. By employing reference images for makeup style transfer, \texttt{DFPP} offers users enhanced flexibility and granular control over desired makeup styles. Furthermore, we demonstrate the effectiveness of our approach in both verification and identification settings, and extend its application to images and videos.

\noindent \textbf{Untrained Neural Network Priors}: While pre-trained generative models have effectively solved a myriad of applications~\cite{shamshad2023evading,asim2019blind,asim2020blind,shamshad2018robust,shamshad2020compressed,shamshad2019adaptive,shamshad2019subsampled,shamshad2020class,shamshad2019deep,xia2022gan}, untrained neural network priors have also demonstrated significant potential in various vision tasks. These untrained  (randomly initialized) neural networks have recently gained traction as effective image priors~\cite{ulyanov2018deep,qayyum2022untrained} for a myriad of visual inverse problems, including denoising~\cite{mataev2019deepred}, super-resolution, inpainting~\cite{schrader2022cnn}, image matching~\cite{hong2021deep}, enhancement~\cite{qayyum2020single,qayyum2022single,shamshad2019subsampled,asim2019patchdip} and scene flow~\cite{li2021neural}. The underpinning idea is that intricate image statistics can be captured by the structure of randomly initialized neural networks, such as CNNs, using the random weights as a parameterization of the resultant output image. While these untrained network priors have found success in various applications~\cite{qayyum2022untrained}, their potential in facial privacy protection remains unexplored. In this work, \textit{for the first time} we show how such priors, when paired with our correspondence module and tailored loss functions, can adeptly safeguard user identity through a natural, yet adversarial, makeup transfer using only a source and makeup reference pair.

\section{Proposed Methodology} 

In this section, we first introduce the protection settings and problem formulation, and then elaborate on the proposed method.

\subsection{Preliminaries} \label{sec:settings}

\textbf{Protection Settings:} Let $\boldsymbol{x} \in \mathcal{X} \subset \mathbb{R}^{n}$ represent a face image, with its normalized feature representation extracted by an FR model as $f(\boldsymbol{x}): \mathcal{X} \rightarrow \mathbb{R}^{d}$. We define a distance metric $\mathcal{D}(\boldsymbol{x}_1,\boldsymbol{x}_2) = D(f(\boldsymbol{x}_1),f(\boldsymbol{x}_2))$ to measure dissimilarity between face images. 
FR systems operate in \emph{verification} and \emph{identification} modes. In verification, two faces are considered identical if $\mathcal{D}(\boldsymbol{x}_1,\boldsymbol{x}_2) \leq \tau$, where $\tau$ is the system threshold. In \emph{closed-set} identification, the system compares a probe image to a gallery, identifying the most similar representation. 
User privacy can be protected by deceiving these malicious FR systems through \emph{impersonation} or \emph{dodging} attacks. Impersonation causes false matches with a target identity, while dodging prevents matches with the same person. These attack strategies apply to both verification and identification scenarios. As attackers can exploit both modes using black-box FR models, effective protection strategies must address all these aspects to comprehensively conceal user identity.

\noindent \textbf{Problem Statement}: Given a source face image $\boldsymbol{x}_s$, our goal is to create a protected face image $\boldsymbol{x}_{p}$ that maximizes $\mathcal{D}(\boldsymbol{x}_{p},\boldsymbol{x}_s)$ for successful dodging and minimizes $\mathcal{D}(\boldsymbol{x}_{p},\boldsymbol{x}_{t})$ for successful impersonation of a target face $\boldsymbol{x}_{t}$, with $\mathcal{O}(\boldsymbol{x}_s) \neq \mathcal{O}(\boldsymbol{x}_{t})$ where $\mathcal{O}$ provides true identity labels. Concurrently, we aim to minimize $\mathcal{H}(\boldsymbol{x}_{p},\boldsymbol{x}_s)$, where $\mathcal{H}$ quantifies the degree of unnaturalness introduced in the protected image $\boldsymbol{x}_{p}$. We formulate this as an optimization problem:
\begin{equation} \label{eq:mai}
\small
    \min{\boldsymbol{x}_p} \mathcal{L}(\boldsymbol{x}_p) = \mathcal{D}(\boldsymbol{x}_p,\boldsymbol{x}_t) - \mathcal{D}(\boldsymbol{x}_p,\boldsymbol{x}_s) 
    \; \text{s.t.} \; \mathcal{H}(\boldsymbol{x}_p,\boldsymbol{x}_s) \leq \epsilon, \notag
\end{equation}
\noindent where $\epsilon$ denotes the bound on the adversarial perturbation. For noise-based approach, $\mathcal{H}(\boldsymbol{x}_p,\boldsymbol{x}_s) = \Vert \boldsymbol{x}_s - \boldsymbol{x}_{p} \Vert_p$, where $\Vert\cdot\Vert_p$ denotes the $L_p$ norm. However, direct enforcement of the perturbation constraint leads to visible artifacts, which affects visual quality and user experience. Constraining the solution search space close to a natural image manifold by imposing an \textit{effective image prior} can produce more realistic images. Note that the distance metric $\mathcal{D}$ is unknown since our goal is to deceive a black-box FR system.

\begin{figure*}[t]
\centering
\vspace{-1em}
\hspace{0.5em}\includegraphics[width=0.95\textwidth]{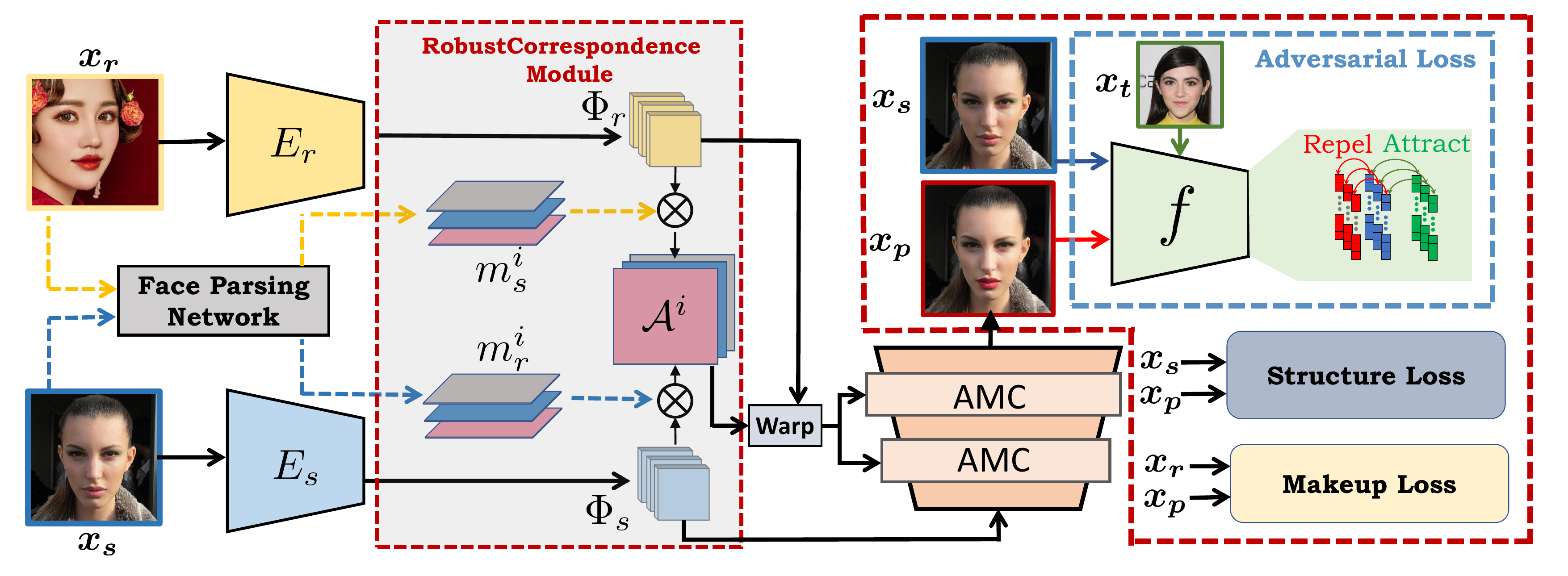}
\caption{Pipeline of the Deep Facial Privacy Prior (\texttt{DFPP}) framework: We employ an encoder-decoder architecture with randomly initialized parameters to adversarially transfer makeup from a reference to a source image, guided by our robust correspondence module. The conditional decoder then aligns the source image to match the reference image features via Adaptive Makeup Conditioning (AMC) layers. Notably, the untrained decoder is test-time finetuned using our structured, makeup, and adversarial losses to effectively protect facial privacy.}
\label{fig:main_figure}
\vspace{-1.1em}
\end{figure*}

\subsection{Deep Facial Privacy Prior (\texttt{DFPP})}  \label{sec:facial_prior}

Our method uniquely leverages the neural network's structure as a prior to generate protected facial images. Unlike previous works that rely on pre-trained models or extensive datasets, we optimize randomly initialized network parameters during inference via gradient descent, capturing an effective facial privacy prior without extensive task-specific training.

\noindent \textbf{Overall Pipeline:} As shown in Fig.~\ref{fig:main_figure}, the \texttt{DFPP} pipeline consists of three key components. First, content encoder $E_s$ and makeup encoder $E_r$ extract multi-scale features from source $\boldsymbol{x}_s$ and reference $\boldsymbol{x}_r$ images, respectively. Next, a region-constrained correspondence module establishes semantic correspondences between $\boldsymbol{x}_s$ and $\boldsymbol{x}_r$ in deep feature space. Finally, a conditional decoder $\mathcal{G}$ synthesizes the protected image $\boldsymbol{x}_p$ using multi-scale features from the correspondence module. The randomly initialized decoder network is optimized at test-time using carefully designed identity preservation, makeup transfer, and adversarial losses. \texttt{DFPP} distinguishes itself from existing makeup-based privacy protection methods by leveraging the network structure itself as a prior.

\subsubsection{Network Architecture}

In this section, we provide details about the architectural components, primarily focusing on the robust correspondence module and the conditional decoder.

\noindent \textbf{Robust Correspondence Module}: 
First, the source $\boldsymbol{x}_s$ and makeup reference $\boldsymbol{x}_r$ images are fed into multi-scale feature extractor networks, $E_s$ and $E_r$, respectively. These networks, pre-trained on ImageNet, extract deep features $\Phi_s = E_s(\boldsymbol{x}_s)$ and $\Phi_r = E_r(\boldsymbol{x}_r)$, both represented in $\mathbb{R}^{C \times H \times W}$, which are then reshaped to $\hat{\Phi}_s$ and $\hat{\Phi}_r$ in $\mathbb{R}^{HW \times C}$. These feature maps contain discriminative information representing the semantics of the inputs. A robust correspondence module then computes a dense semantic correspondence matrix $\mathcal{A} \in \mathbb{R}^{HW \times HW}$, which represents how pixels in $\boldsymbol{x}_s$ are morphed from pixels in $\boldsymbol{x}_r$. To avoid artifacts, makeup should be transferred between pixels with similar relative positions (\textit{e.g.}, lips to lips), reflected by high correlation values $\mathcal{A}(u,v)$ between these pixels.

A naive way to find the correlation (attention) matrix is to compare the similarity between the feature maps $\Phi_s$ and $\Phi_r$ as $\mathcal{A}(u,v) = \frac{\hat{\Phi}_s(u)^{T}\hat{\Phi}_r(v)}{\Vert \hat{\Phi}_s(u) \Vert \Vert \hat{\Phi}_r(v) \Vert},$
where $\hat{\Phi}_s(u) \in \mathbb{R}^{C \times 1}$ and $\hat{\Phi}_r(v) \in \mathbb{R}^{C \times 1}$ represent the channel-wise centralized features at position $u$ and $v$, respectively~\cite{he2018deep}.
Next, the reference features $\hat{\Phi}_r$ are warped to the source features $\hat{\Phi}_s$ according to $\mathcal{A}$, creating spatially aligned reference-to-source features $\hat{\Phi}_{s \leftarrow r}$ as $\hat{\Phi}_{s \leftarrow r}(u) = \sum_v \text{softmax} (\alpha \mathcal{A}(u,v)) \hat{\Phi}_r(v)$, where $\alpha$ is the temperature parameter to control the sharpness of softmax across $v$.  However, this naive approach often yields poor results due to false matches in semantic correspondence, especially in the presence of adversarial toxicity. In our case, this issue is particularly severe because we only have a single source and reference image to establish correspondence.

To address false matches due to adversarial toxicity, we propose spatially constraining semantic correspondences among facial regions of $\boldsymbol{x}_s$ and $\boldsymbol{x}_r$ in deep feature space, using facial parsing masks as guidance. Let $m_s^i$ and $m_r^i$ denote face parsing masks for $\boldsymbol{x}_s$ and $\boldsymbol{x}_r$, where $i \in { \text{eye, lip, skin}}$. Region-constrained deep features are obtained as $\Phi_s^{i} = \Phi_s \odot m_s^i$ and $\Phi_r^{i} = \Phi_r \odot m_r^i$, where $\odot$ is element-wise multiplication. Robust correspondences are established via correlation matrices:
\begin{equation}
\small
    \mathcal{A}^{i}(u,v) = \frac{\hat{\Phi}^i_s(u)^{T}\hat{\Phi}^i_r(v)}{\Vert \hat{\Phi}^i_s(u) \Vert \Vert \hat{\Phi}^i_r(v) \Vert}. \label{eq:reg_attention}
\end{equation}

\noindent Using $\mathcal{A}^i$, we spatially align the region constrained makeup features with the corresponding source features via warping  and concatenate them to obtain the final warped features $\hat{\Phi}^{conc.}_s $ after passing through a $1 \times 1$ convolution layer.

\noindent \textbf{Conditional Decoder}: Guided by the final warped makeup features $\hat{\Phi}^{conc.}_{s \leftarrow r}$ and source features $\Phi_{s}$, the conditional decoder $\mathcal{G}_\theta$ generates protected image $\boldsymbol{x}_p$ that respects the spatial semantic structure of $\boldsymbol{x}_s$ and makeup style of $\boldsymbol{x}_r$ as: $\boldsymbol{x}_p = \mathcal{G}_\theta (\hat{\Phi}^{conc.}_{s \leftarrow r},\Phi_{s})$.  
In order to effectively use the warped final makeup features to guide the generation, and to better preserve the makeup style information, we use spatially-adaptive denormalization (SPADE)~\cite{park2019semantic} in $\mathcal{G}_\theta$. Specifically, we progressively inject $\Phi_{s}$ at different scales to modulate the activation functions of the SPADE block in $\mathcal{G}_\theta$. Unlike prior works that employ fixed decoder parameters $\theta$ obtained after an intensive training process on a large makeup dataset, we initialize the parameters of the conditional decoder randomly and optimize them during test-time to effectively capture the source-reference pair-specific prior guided by explicit content, makeup, and adversarial objective functions.

\subsubsection{Loss Function}
\label{subsec:Objective Functions}
Our overall objective function focuses on three critical aspects of the protected image $\boldsymbol{x}_p$: the \textit{structure loss} ensures the preservation of the human-perceived identity from $\boldsymbol{x}_s$; the \textit{makeup loss} robustly transfers face makeup of $\boldsymbol{x}_r$ to the relevant semantic regions of $\boldsymbol{x}_s$, and \textit{adversarial loss} generates effective adversarial perturbations to evade black-box FR models.

\noindent \underline{\textbf{Structure Loss}}:
Existing makeup transfer methods typically rely on perceptual loss in the VGG feature space to preserve the identity of the source face~\cite{nguyen2021lipstick}. However, this loss may suffer from two issues in the presence of adversarial toxicity. \textit{Firstly}, it can cause distortion of the facial attributes of the source image. \textit{Secondly}, a trade-off between preserving the original identity of $\boldsymbol{x}_s$ and maintaining a high protection success rate may arise due to conflicting objectives. 
Inspired by recent findings~\cite{tumanyan2022splicing,bar2022text2live} that the deep features in the multi-head self-attention (MSA) layer of the pre-trained DINO-ViT~\cite{caron2021emerging} contain crucial structural information, we introduce a loss function that effectively maintains the structural consistency between the $\boldsymbol{x}_s$ and $\boldsymbol{x}_p$. Specifically, we define the structure loss as a difference in self-similarity $S(.)$ of the keys extracted from the attention module at the deepest transformer layer. The structure loss can be expressed as:
\begin{equation}
\small
\mathcal{L}_{\text{struc}}(\boldsymbol{x}_{s},\boldsymbol{x}_p) = \|S^l(\boldsymbol{x}_{s})-S^l(\boldsymbol{x}_p)\|_F, 
\end{equation}
where $[S^l(\boldsymbol{x})]_{i,j} = \cos(k^l_i(\boldsymbol{x}),k^l_j(\boldsymbol{x}))$. Here,  $k^l_i(\boldsymbol{x})$ and $k^l_j(\boldsymbol{x})$ represents $i^{th}$ and $j^{th}$ keys in the $l^{th}$ MSA layer of pre-trained ViT with image $\boldsymbol{x}$ and `cos' denotes cosine similarity. 
We apply this loss in a patch-wise contrastive manner~\cite{park2020contrastive,jung2022exploring} to ensure that keys at the same positions have closer distances while maximizing distances between keys at different positions. This approach effectively preserves the source image's structure (identity) during adversarial optimization.

\noindent \underline{\textbf{Robust Makeup Transfer Loss}}:
The primary objective of robust makeup transfer loss is  to achieve adversarial makeup transfer between corresponding regions of $\boldsymbol{x}_s$ and $\boldsymbol{x}_r$, while maintaining global coherence and preventing artifacts in non-makeup areas (\textit{e.g.}, teeth, hair, background). To address challenges posed by adversarial toxicity during optimization, we employ two main components: a \textbf{\textit{Histogram Matching Loss}} that matches color histograms in corresponding regions of source and reference makeup images, and a \textbf{\textit{Global Loss}} that maintains the overall tone of the reference makeup style.

The \textbf{\textit{Histogram Matching}} (HM) Loss applies color histogram matching to corresponding facial regions (skin, lips, and eyes) using face parsing masks. It aims to equalize the color distribution between regions of $\boldsymbol{x}p$ and $\boldsymbol{x}r$~\cite{li2018beautygan,gu2019mask}. Consequently, the HM loss is formulated as the weighted sum of the corresponding local regional losses and can be expressed as $\mathcal{L}_{HM} = \lambda_{\text{lips}} \mathcal{L}_{\text{lips}}+\lambda_{\text{eyes}} \mathcal{L}_{\text{eyes}}+\lambda_{\text{skin}} \mathcal{L}_{\text{skin}}$,
where $\lambda_{\text{lips}}$, $\lambda_{\text{eyes}}$, and $\lambda_{\text{skin}}$ are hyperparameters. Specifically, each loss
item is a local histogram loss, which can be written as:
\begin{equation}
\small
    \mathcal{L}_{i} = \Vert \boldsymbol{x}_p\odot m^{i}_s - \text{HM}(\boldsymbol{x}_s\odot m^{i}_s,\boldsymbol{x}_r\odot m^{i}_r) \Vert,
\end{equation}
where $\odot$ is pixel-wise multiplication, and $m_s$ and $m_r$ are face parsing masks. The resulting histogram-matched regions form a pseudo-ground truth, providing coarse guidance during test-time adversarial optimization. While this discards spatial information, it offers sufficient guidance for makeup color transfer, which is crucial in the presence of adversarial toxicity.

The second component of our robust makeup transfer loss is the \textbf{\textit{Global Loss}}, which ensures faithful transfer of global makeup elements from $\boldsymbol{x}_r$ to $\boldsymbol{x}_s$. Defined in a patch-wise and multi-scale manner for effective photorealistic transfer, it is expressed as:
\begin{equation}
\small
    \mathcal{L}_{\text{glob}} =  \sum_l\sum_u \Vert \Psi ((\Phi^{conc.}_{s \leftarrow r}(u)) -  \Psi (\Phi_r(NN(u))) \Vert _F,
\end{equation}
where $\Psi(.)$ extracts local patches, and $NN(u)$ is the index of the nearest patch in $\Phi_r$ to $\Psi (\Phi^{conc.}{s \leftarrow r}(u))$, found using a cross-correlation matrix.  This matrix establishes the similarity between patches of source features and makeup face features. Our overall robust makeup loss combines the histogram and global losses: $\mathcal{L}{\text{makeup}} = \mathcal{L}_{\text{HM}}+\mathcal{L}_{\text{glob}}$.

\noindent \underline{\textbf{Adversarial Loss}}:  We optimize the randomly initialized parameters of the untrained conditional decoder $\mathcal{G}_\theta$ to find a protected face $\boldsymbol{x}_p$ whose feature representation is close to the target image $\boldsymbol{x}_t$ and far from the original image $\boldsymbol{x}_s$. This adversarial loss is expressed as:
\begin{equation} \label{eq:reform}
\small
    \mathcal{L_{\text{adv}}} = \mathcal{D}(\mathcal{G}_\theta (\Phi_{s \leftarrow r},\Phi_{s}),\boldsymbol{x}_t) - \mathcal{D}(\mathcal{G}_\theta (\Phi_{s \leftarrow r},\Phi_{s}),\boldsymbol{x}_s), 
\end{equation}
where $\mathcal{D}(\boldsymbol{x}_1,\boldsymbol{x}_2) = 1-\text{cos}[f(\boldsymbol{x}_1),f(\boldsymbol{x}_2))]$ is the cosine distance. In the black-box setting, we optimize on an ensemble of white-box surrogate models to craft transferable attacks that mimic the unknown FR model's decision boundary.

Finally, combining all the loss functions, we have $\mathcal{L}_{\text{total}} =  \lambda_{\text{struc}} \mathcal{L}_{\text{struc}} +  \lambda_{\text{makeup}} \mathcal{L}_{\text{cycle}}+ \lambda_{\text{adv}} \mathcal{L}_{\text{adv}}$,
where $\lambda$ terms are hyperparameters. $\mathcal{L}_{\text{struc}}$ preserves the human perceived identity of the image, $\mathcal{L}_{\text{makeup}}$ ensures faithful makeup transfer in relevant regions, and $\mathcal{L}_{\text{adv}}$ accounts for the adversarial objective to fool malicious FR models.

\section{Experiments} \label{sec:experiments}

\noindent \textbf{Implementation details:} 
We use the Adam optimizer~\cite{kingma2014adam} ($\beta_1=0.9$, $\beta_2=0.999$, learning rate $2 \times 10^{-4}$) for 450 iterations on A100 GPUs. Face parsing is done with BiSeNet~\cite{yu2018bisenet}, followed by mask smoothing to ensure a seamless transition around the edges~\cite{masi2020towards}.

\noindent \textbf{Hyperparameters:} \label{sec:hyp}
The components of our loss functions are weighted as follows: structure loss ($\lambda_\text{ViT}=0.001$), makeup loss ($\lambda_\text{hist}=0.8$, $\lambda_\text{glob}=0.2$), and adversarial loss ($\lambda_\text{adv}=0.003$).

\noindent \textbf{Baseline methods:} \label{sec:baseline}
We compare \texttt{DFPP} with recently proposed noise-based and makeup-based privacy protection approaches. Noise-based methods include PGD~\cite{DBLP:conf/iclr/MadryMSTV18}, MI-FGSM~\cite{DBLP:conf/cvpr/DongLPS0HL18}, TI-DIM~\cite{DBLP:conf/cvpr/DongPSZ19}, and TIP-IM~\cite{yang2021towards}, and makeup-based approaches include Adv-Makeup~\cite{DBLP:conf/ijcai/YinWYGKDLL21} and AMT-GAN~\cite{hu2022protecting}. TIP-IM also incorporate multi-target objective in its optimization to find the optimal target image among multiple targets.  For fair comparison with AMT-GAN~\cite{hu2022protecting}, we use TIP-IM's single-target variant in main experiments. We also present multi-target results to demonstrate \texttt{DFPP}'s effectiveness in such scenarios. In our main experiments, we do not include methods requiring pre-trained high-quality generators like CLIP2Protect~\cite{shamshad2023clip2protect} and DiffAM\footnote{\small The code of DiffAM is not publicly available at the time of submission.}~\cite{sun2024diffam}. Additionally, CLIP2Protect is text-based, while our approach is image-based, making direct comparison less meaningful. However, we do compare with CLIP2Protect in a separate analysis to demonstrate that \texttt{DFPP} is less gender-biased.

\noindent \textbf{Datasets:}  
For \underline{\textit{face verification}}, we use CelebA-HQ~\cite{karras2018progressive} and LADN~\cite{gu2019ladn} for the impersonation attack. We follow the settings of AMT-GAN~\cite{hu2022protecting} and select a subset of 1,000 images from CelebA-HQ, reporting average results over the 4 target identities provided by~\cite{hu2022protecting}. Similarly, for LADN, we divide the 332 images into 4 groups, where images in each group aim to impersonate the target identities provided by \cite{hu2022protecting}. For the dodging attack, we use CelebA-HQ~\cite{karras2018progressive} and LFW datasets~\cite{parkhi2015deep} by selecting 500 subjects at random, where each subject has a pair of faces. 
For \textit{\underline{face identification}}, we use CelebA-HQ~\cite{karras2018progressive} and LFW~\cite{huang2008labeled} as our evaluation set for both impersonation and dodging attacks. For both datasets, we randomly select 500 subjects, each with a pair of faces. We assign one image in the pair to the gallery set and the other to the probe set. Both impersonation and dodging attacks are performed on the probe set. For impersonation, we insert 4 target identities provided by~\cite{hu2022protecting} into the gallery set. For all experiments, we use 10 reference makeup images provided by~\cite{hu2022protecting}.
Regarding pre-processing, we use MTCNN~\cite{zhang2016joint} to detect, crop and align the face image before giving it as input to FR models. 


\noindent \textbf{Target Models:} We evaluate \texttt{DFPP}'s effectiveness against four black-box FR models: IRSE50~\cite{hu2018squeeze}, IR152~\cite{deng2019arcface}, FaceNet~\cite{schroff2015facenet}, and MobileFace~\cite{chen2018mobilefacenets}. All input images are pre-processed using MTCNN~\cite{zhang2016joint} for face detection and alignment. We also test \texttt{DFPP} against commercial APIs: Face++ and Tencent Yunshentu.

\noindent \textbf{Evaluation metrics:} We evaluate \texttt{DFPP} using the Protection Success Rate (PSR)\cite{yang2021towards}, which measures the fraction of protected faces misclassified by FR models, employing thresholding for verification and a closed-set strategy for identification. For face identification, we also use Rank-N Targeted Identity Success Rate (Rank-N-T), indicating the target image appears at least once in the top N gallery candidates, and Rank-N Untargeted Identity Success Rate (Rank-N-U), where the top N candidates exclude the original image's identity. To assess the realism of protected images, we report FID\cite{heusel2017gans} scores. For commercial APIs, we directly report the confidence scores returned by the respective servers.
\begin{table}[t]
\caption{\small Protection success rate (PSR \%) of black-box impersonation attack under face verification task where for each column the other three FR models are used as surrogates.  }
\centering
\label{table:verification_impersonation}
\vspace{-2mm}
\setlength{\tabcolsep}{2pt}
\scalebox{0.8}{
\begin{tabular}{l || c | c | c | c || c | c | c | c || c }
\toprule[0.15em]
\rowcolor{mygray} \textbf{Method} & \multicolumn{4}{c||}{\textbf{CelebA-HQ}}&\multicolumn{4}{c||}{\textbf{LADN-Dataset}}&\multicolumn{1}{c}{\textbf{Avg.}} \\
\rowcolor{mygray}  & IRSE50 & IR152 & FaceNet& MobFace & IRSE50 & IR152 & FaceNet& MobFace &  \\
\midrule[0.15em]
Clean & 7.29  & 3.80  & 1.08  & 12.68  & 2.71  & 3.61  & 0.60  & 5.11  & 4.61   \\
PGD~\cite{DBLP:conf/iclr/MadryMSTV18}&  36.87  & 20.68 & 1.85    & 43.99      & 40.09   & 19.59  & 3.82     & 41.09 & 25.60 \\
MI-FGSM~\cite{DBLP:conf/cvpr/DongLPS0HL18} & 45.79  & 25.03 & 2.58     & 45.85      & 48.90    & 25.57  & 6.31     & 45.01 & 30.63 \\
TI-DIM~\cite{DBLP:conf/cvpr/DongPSZ19} & 63.63&36.17& 15.30&57.12& 56.36&34.18& 22.11&48.30& 41.64 \\
$\text{Adv-Makeup}$~\cite{DBLP:conf/ijcai/YinWYGKDLL21} & 21.95  & 9.48  & 1.37    & 22.00      & 29.64   & 10.03  & 0.97     & 22.38& 14.72  \\
$\text{TIP-IM}$~\cite{yang2021towards}  & 54.40 & 37.23 & 40.74& 48.72 & 65.89 & 43.57 & \textbf{63.50} & 46.48 & 50.06 \\
$\text{AMT-GAN}$~\cite{hu2022protecting}  & 76.96 & 35.13 & 16.62 & 50.71 & 89.64 & 49.12 & {32.13} & {72.43} & 52.84  \\ 
\midrule
\rowcolor{cyan!20}$ {\text{DFPP}} \; {\text{(Ours)}}$  & \textbf{78.25} & \textbf{41.25}  & \textbf{40.86}  & \textbf{69.34} & \textbf{90.27} & \textbf{51.66} & {49.91} & \textbf{77.14} & \textbf{62.34}   \\
\bottomrule[0.1em]
\end{tabular}}
\vspace{-1em}
\end{table}


\begin{wrapfigure}{r}{0.65\textwidth}
\vspace{-4em}
\begin{minipage}{\linewidth}
\captionof{table}{\small Protection success rate (PSR \%) of \textit{black-box} dodging (top) and impersonation  (bottom) attacks under the face identification task for LFW dataset~\cite{huang2008labeled} where for each column the other three \texttt{FR} systems are used as surrogates.
R1-U: Rank-1-Untargeted, R5-U: Rank-5-Untargeted,  R1-T: Rank-1-Targeted, R5-T: Rank-5-Targeted. }
\label{table:identificationlfw}
\vspace{0mm}
\setlength{\tabcolsep}{1.8pt}
\scalebox{0.62}{
\begin{tabular}{l || c c || c c || c c || c c || c c }
\toprule[0.15em]
\rowcolor{mygray} \textbf{Method} & \multicolumn{2}{c||}{\textbf{IRSE50}}&\multicolumn{2}{c||}{\textbf{IR152}}&\multicolumn{2}{c||}{\textbf{FaceNet}}&\multicolumn{2}{c||}{\textbf{MobileFace}}&\multicolumn{2}{c}{\textbf{Average}} \\
\rowcolor{mygray}  & R1-U & R5-U & R1-U & R5-U & R1-U & R5-U & R1-U & R5-U & R1-U & R5-U  \\
\midrule[0.15em]
MI-FGSM~\cite{DBLP:conf/cvpr/DongLPS0HL18}  & 70.2 & 42.6 & 58.4 & 41.8 & 59.2 & 34.0 & 68.0 & 47.2 & 63.9 &41.4  \\
TI-DIM~\cite{DBLP:conf/cvpr/DongPSZ19} & 79.0 &51.2 & 67.4 & 54.0 & 74.4 & 52.0 & 79.2 & 61.6 & 75.0&54.7 \\
$\text{TIP-IM}$~\cite{yang2021towards} & 81.4 & 52.2& 71.8 & 54.6 & 76.0 & 49.8 & 82.2 & 63.0 & 77.8&54.9 \\
\midrule
\rowcolor{cyan!20}$ {\text{DFPP}} \; {\text{(Ours)}}$  & \textbf{82.2} &\textbf{55.6} & \textbf{73.0} & \textbf{55.4} & \textbf{80.8} & \textbf{53.4} & \textbf{84.2} & \textbf{66.6} & \textbf{80.05} & \textbf{57.75}  \\
\midrule[0.15em]
\rowcolor{mygray}  & R1-T & R5-T & R1-T & R5-T & R1-T & R5-T & R1-T & R5-T & R1-T & R5-T  \\
\midrule[0.15em]
MI-FGSM~\cite{DBLP:conf/cvpr/DongLPS0HL18}  & 4.0 & 10.2 & 3.2 & 14.2 & 9.0 & 18.8 & 8.4 & 22.4& 6.15 & 16.4\\
TI-DIM~\cite{DBLP:conf/cvpr/DongPSZ19} & 4.0 &13.6 & 7.8& 19.6 & 18.0 & 32.8 & 21.6 & 39.0 & 12.85& 26.25 \\
$\text{TIP-IM}$~\cite{yang2021towards} & 8.0 &28.2& 11.6 & 31.2 & 25.2 & \textbf{56.8} & 34.0 & 51.4 & 19.7& 41.9\\
\midrule
\rowcolor{cyan!20}$ {\text{DFPP}} \; {\text{(Ours)}}$  & \textbf{10.6} &\textbf{33.2} &\textbf{12.8} & \textbf{37.2} & \textbf{26.0} &  52.8 & \textbf{36.6} &\textbf{58.2} &\textbf{21.50} & \textbf{45.35}  \\
\bottomrule[0.1em]
\end{tabular}}
\vspace{-3em}
\end{minipage}
\end{wrapfigure}

\subsection{Experimental Results} \label{sec:experiments}

We generate protected images using three surrogate FR models to mimic the decision boundary of the fourth black-box FR model, employing 10 reference makeup images as per AMT-GAN~\cite{hu2022protecting}. For face verification, we set the system threshold at 0.01 false match rate for each FR model: IRSE50 (0.241), IR152 (0.167), MobileFace (0.302), and FaceNet (0.409). Tab. \ref{table:verification_impersonation} presents quantitative results for impersonation attacks in face verification, demonstrating \texttt{DFPP}'s superior performance with average absolute PSR gains of 10$\%$ and 12$\%$ over AMT-GAN~\cite{hu2022protecting} and the noise-based method~\cite{yang2021towards}, respectively.
The PSR values for dodging and impersonation attacks under the face identification task on the LFW dataset are presented in Tab. \ref{table:identificationlfw}, where \texttt{DFPP} demonstrates superior performance compared to recent methods at both \textit{Rank-1} and \textit{Rank-5} settings. For this experiment, we randomly select 500 subjects, each having a pair of faces. We assign one image from each pair to the gallery set, and the other to the probe set. Both impersonation and dodging attacks are conducted on the probe set. We excluded AMT-GAN and Adv-Makeup from both tables, as they are specifically trained for the face verification task.
Results for dodging attacks under face verification and impersonation attacks under the face identification are provided in the supplementary material.
Qualitative results in Fig.~\ref{fig:qual_2} also demonstrate \texttt{DFPP}'s superiority in generating realistic protected faces. Unlike TIP-IM's noise artifacts and AMT-GAN's unrealistic makeup effects, \texttt{DFPP} produces natural-looking faces that faithfully replicate the reference image's makeup style.

\begin{table}[t]
\centering
\begin{minipage}{.4\linewidth}
  \caption{\small FID and PSR comparison. PSR Gain is absolute gain in PSR relative to Adv-Makeup.}
  \label{table:fid_}
  \centering
  \setlength{\tabcolsep}{6.0pt}
  \scalebox{0.7}{
  \begin{tabular}{l | c  |c }
  \toprule[0.15em]
  \rowcolor{mygray} \textbf{Method} & FID $\downarrow$  & PSR Gain $\uparrow$ \\
  \midrule[0.15em]
  Adv-Makeup~\cite{DBLP:conf/ijcai/YinWYGKDLL21} & 4.23  &  0    \\
  TIP-IM~\cite{yang2021towards}  & 38.73 &  35.34 \\
  AMT-GAN~\cite{hu2022protecting}  & 34.44  & 38.12\\
  CLIP2Protect~\cite{shamshad2023clip2protect}  & 26.62  & 50.18\\
  \midrule
  \rowcolor{cyan!20} DFPP (Ours) & 29.81 & 47.62 \\
  \bottomrule[0.1em]
  \end{tabular}}
\end{minipage}%
\hspace{0.02\linewidth}
\begin{minipage}{.54\linewidth}
  \caption{\small PSR comparison on male and female faces (MobileFace as black-box). \texttt{DFPP} provides balanced protection (female/male ratio $\approx$ 1).}
  \label{tab:bias_}
  \centering
  \setlength{\tabcolsep}{3pt}
  \scalebox{0.95}{%
  \begin{tabular}{l  |c|c|c|c}
  \toprule[0.15em]
  \rowcolor{mygray} Methods  & Images & Male & Female & Ratio \\
  \midrule[0.15em]
  AMT-GAN~\cite{hu2022protecting} &  1000& 511& 722& 1.41 \\
  CLIP2Protect~\cite{shamshad2023clip2protect} &  829& 904& 722& 1.09 \\
  \rowcolor{cyan!20} Ours  & 1000 & 802 & 817 & 1.02 \\
  \bottomrule[0.1em]
  \end{tabular}}
\end{minipage} 
\vspace{-1.4em}
\end{table}

\begin{wrapfigure}{r}{0.42\textwidth} 
\vspace{-3em}
  \begin{center}
    \includegraphics[clip, trim=0cm 0cm 0cm 0cm,width=0.42\textwidth]{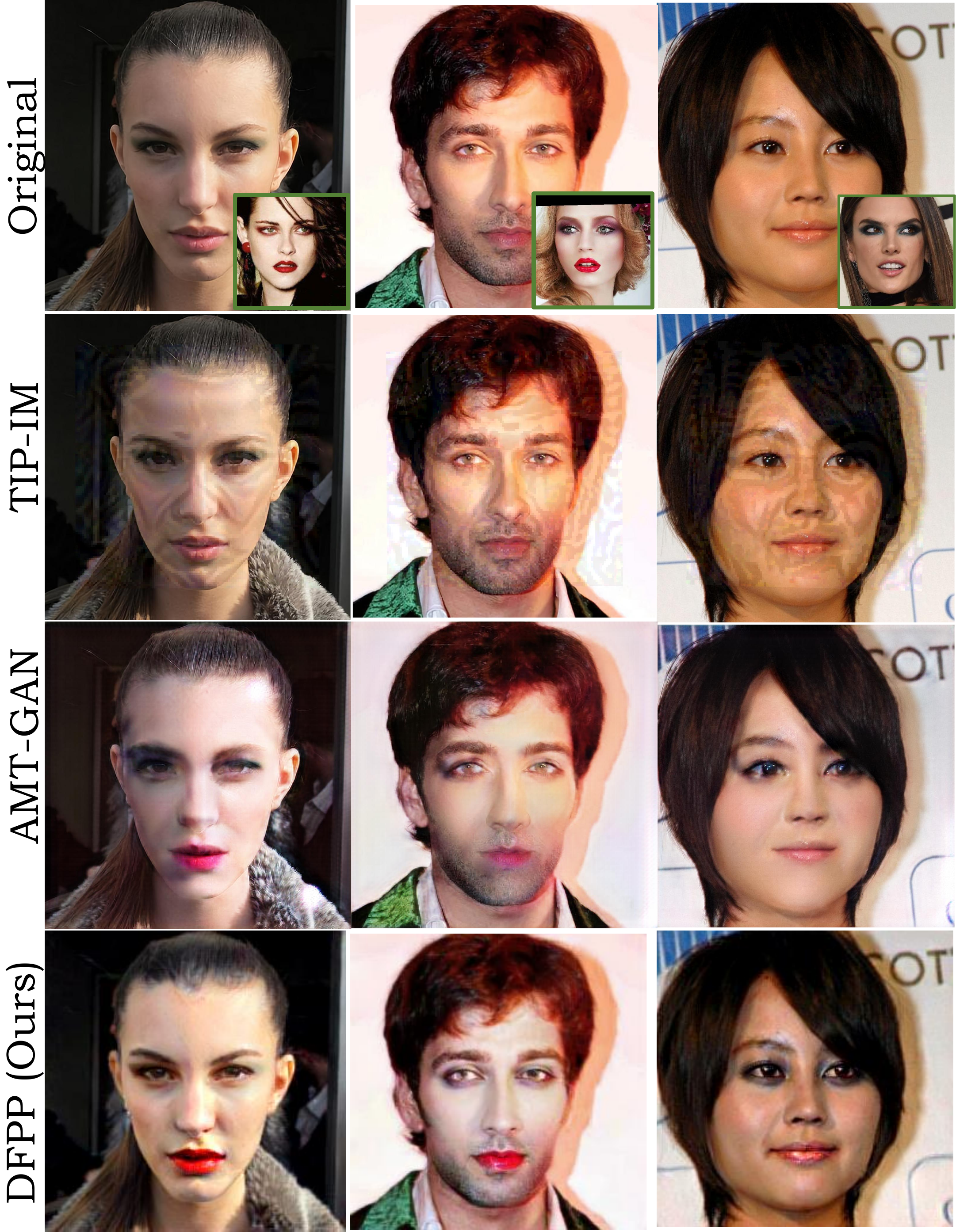}
    \caption{\small Qualitative comparison of \texttt{DPFF} with TIP-IM~\cite{yang2021towards} and AMT-GAN~\cite{hu2022protecting} approaches. \texttt{DPFF} generates naturalistic images that maintain the human-perceived identity of the original, while faithfully transferring the makeup from the reference image (shown in the top row of the bottom corner).\vspace{-3em}}
    \label{fig:qual_2}
  \end{center}
\end{wrapfigure}


\noindent \textbf{FID Score:} \noindent \textbf{FID Score:} Tab. \ref{table:fid_} shows FID scores (lower is better) for makeup-based methods. \texttt{DFPP} achieves lower FID scores and higher PSR than TIP-IM and AMT-GAN, balancing protection and naturalness. While Adv-Makeup has the lowest FID, its PSR is lower due to limited eye-area application. Notably, \texttt{DFPP}'s results are comparable to CLIP2Protect~\cite{shamshad2023clip2protect}, despite the latter's high-quality pre-trained generator, indicating the strong image prior imposed by untrained neural networks.

\noindent \textbf{Dataset Bias:}
We evaluate gender bias using 1,000 male and 1,000 female faces from CelebA-HQ, generating protected faces with \texttt{DFPP} and MobileFace as the black-box model for impersonation in face verification. Tab.~\ref{tab:bias_} shows that, in contrast to AMT-GAN\cite{hu2022protecting} and CLIP2Protect~\cite{shamshad2023clip2protect}, \texttt{DFPP}'s PSR is not significantly affected by gender, providing balanced protection (female/male ratio close to 1) for both male and female faces.

\begin{wrapfigure}{r}{0.5\textwidth} 
\vspace{-6.5em}
  \begin{center}
    \includegraphics[clip, trim=0cm 0.2cm 0cm 0cm,width=0.5\textwidth]{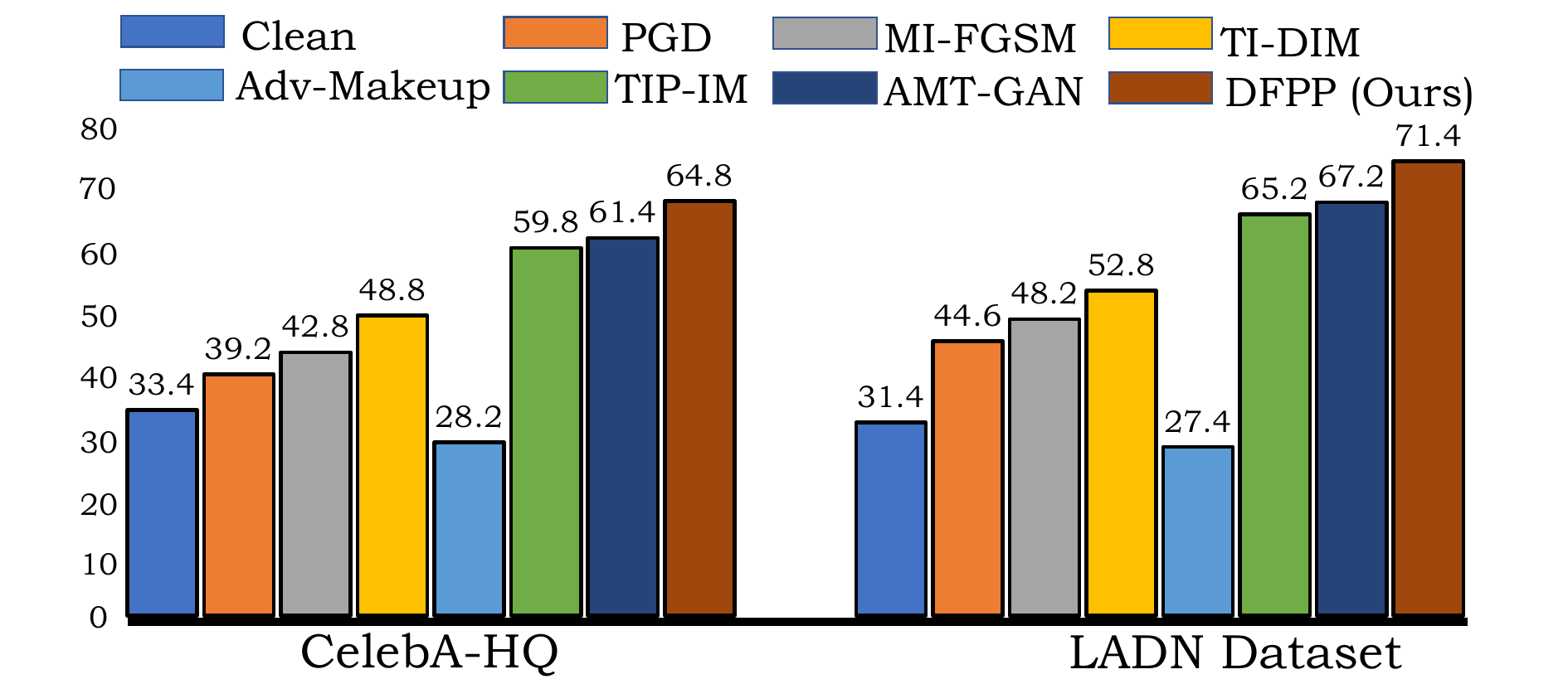}
    \caption{\small Average confidence scores (where higher scores are preferable) from commercial API, Face++ for impersonation attacks within the face verification framework. The \texttt{DFPP} method consistently outperforms these approaches.\vspace{-3.5em}}
    \label{fig:face++}
  \end{center}
\end{wrapfigure}
\subsection{Effectiveness against Commercial APIs} \label{sec:real-world}
We evaluate \texttt{DFPP}'s performance against commercial APIs (Face++ and Tencent Yunshentu) in verification mode for impersonation. These APIs return confidence scores (0-100) to measure image similarity, with higher scores indicating greater similarity. This test simulates real-world scenarios, as the training data and model parameters of these proprietary FR models are undisclosed. We protect 100 faces from CelebA-HQ using \texttt{DFPP} and baseline methods. Fig. \ref{fig:face++} shows the average confidence scores returned by Face++, demonstrating \texttt{DFPP}'s superior PSR compared to baselines. The results for Tencent Yunshentu are provided in the supplementary material.

\subsection{Extension to Videos}

\begin{wrapfigure}{r}{0.6\textwidth} 
\vspace{-6.5em}
  \begin{center}
    \includegraphics[clip, trim=0cm 0.0cm 17.9cm 0cm,width=0.53\textwidth]{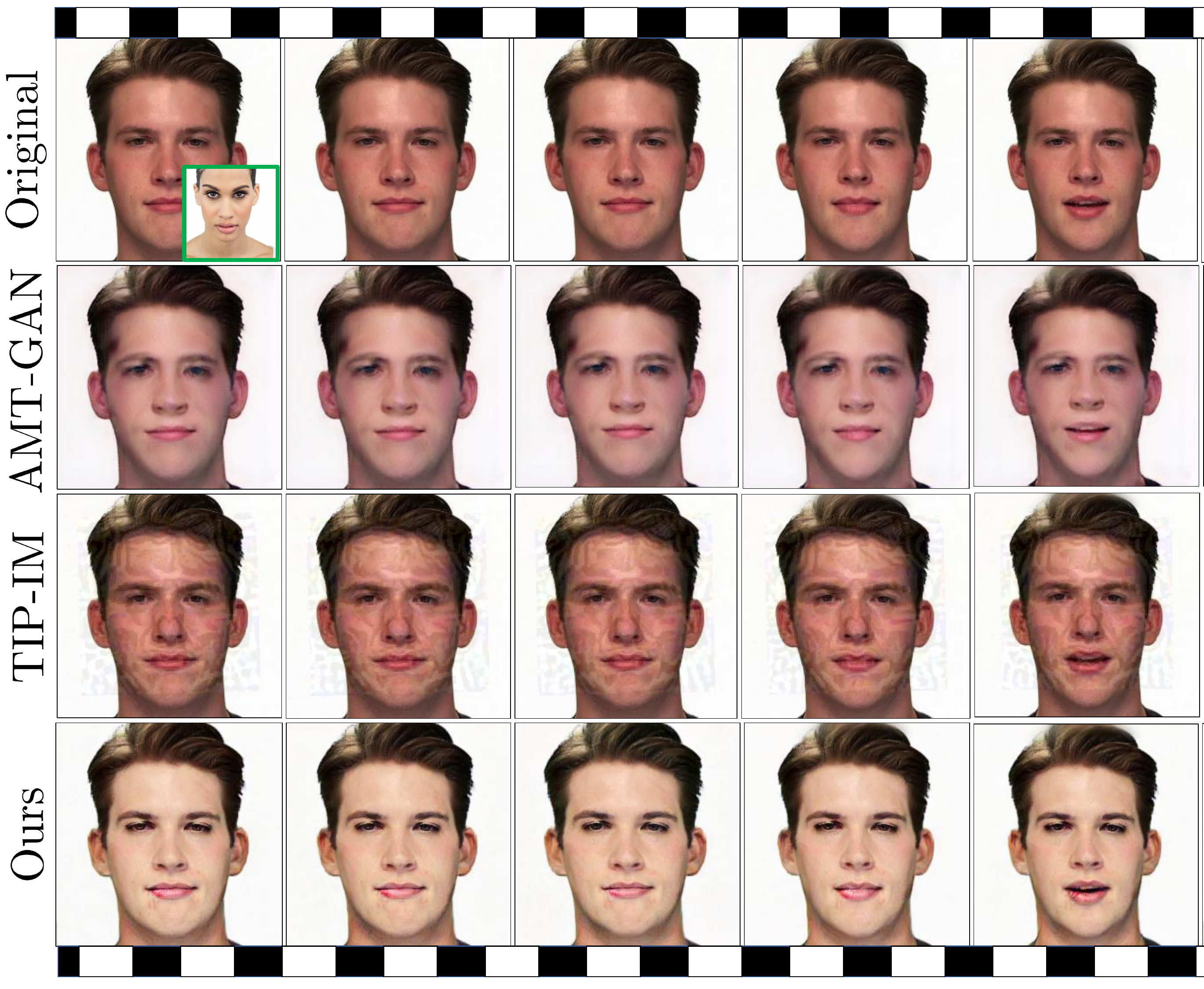}
    \caption{\footnotesize Qualitative results on videos produced by AMT-GAN~\cite{hu2022protecting}, TIP-IM~\cite{yang2021towards}, and our approach for black-box impersonation attacks in the face verification framework. The reference makeup image is highlighted in the green box. Our approach yields more natural outputs that accurately mimic the makeup style of the reference image. \vspace{-2.8em}}
    \label{fig:videos}
  \end{center}
\end{wrapfigure}

We extend our approach to videos by leveraging temporal information. Specifically, for each subsequent frame, we initialize the decoder parameters using those optimized for the preceding frame. This strategy provides an advantageous initialization for our optimization, facilitating faster convergence. Evaluations on 10 randomly chosen videos from the RAVDESS dataset~\cite{livingstone2018ryerson} indicate that \texttt{DFPP} outperforms AMT-GAN, achieving an absolute improvement of $3.2$ in FID score, all the while requiring $10\times$ fewer iterations compared to its image-centric counterpart.  Fig.~\ref{fig:videos} illustrates qualitative results, demonstrating the superior naturalness of our method and the adherence to the reference makeup style.

\subsection{Ablation Studies} \label{sec:ablations}
In this section, we conduct ablation studies to assess the significance of individual components within our overall framework.

\begin{figure}[t]
\centering
\begin{minipage}{.48\linewidth}
  \centering
  \includegraphics[width=\linewidth]{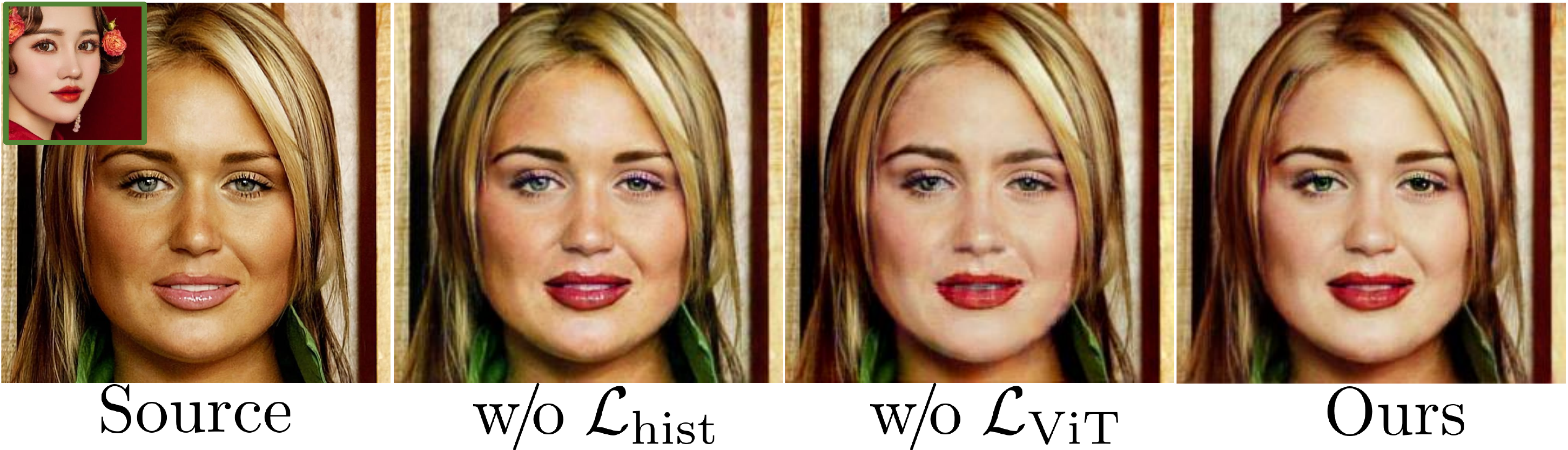}
  \caption{Ablation study on histogram and ViT structural losses. Removing the histogram loss results in suboptimal makeup color transfer, while omitting the ViT loss impairs source identity preservation. Our method effectively maintains both the reference makeup style and source identity.} 
  \label{fig:ablation_loss}
\end{minipage}%
\hspace{0.02\linewidth}%
\begin{minipage}{.48\linewidth}
  \centering
  \includegraphics[width=\linewidth]{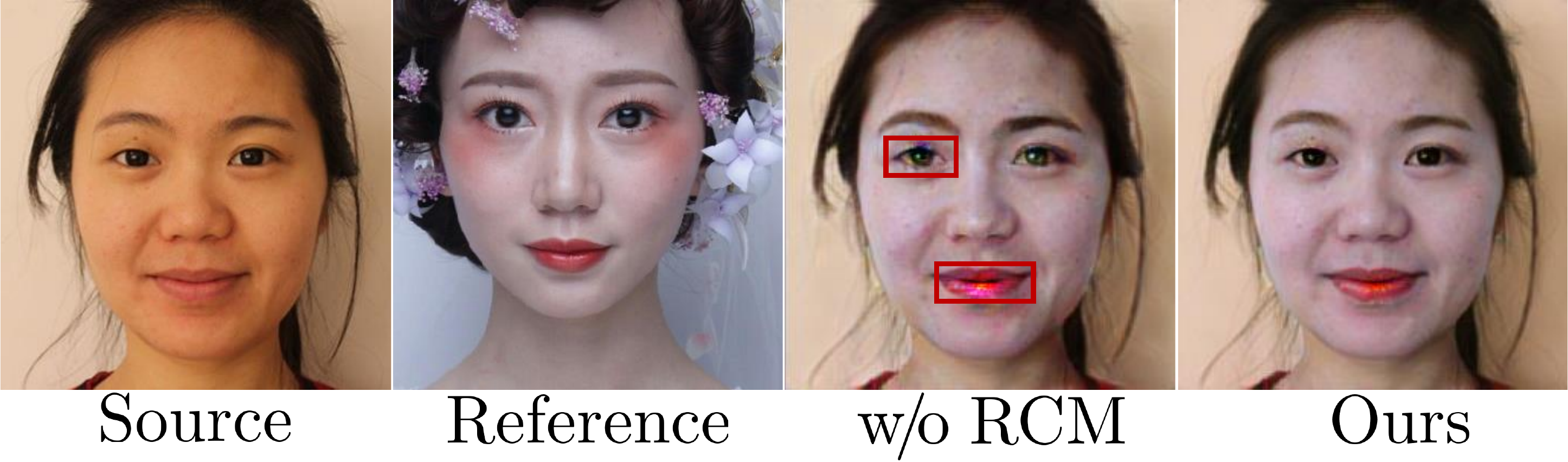}
  \caption{\small Ablation for robust correspondence module (RCM). Without RCM, results show makeup artifacts from adversarial toxicity, highlighting the module's importance in ensuring faithful makeup transfer between corresponding regions of the source and reference images.}
  \label{fig:robust_module}
\end{minipage}
\end{figure}

\noindent \textbf{Loss functions.}
We dissect the performance of individual loss components, both qualitatively and quantitatively. The results presented in Fig.~\ref{fig:ablation_loss} highlight the significance of each loss components. Specifically, omitting the histogram loss leads to an imperfect transfer of makeup color from the reference makeup to the source image. On the other hand, the absence of the ViT structure loss results in a subtle alteration in the identity of the source image compared to when our full objective is employed.  We further provide quantitative analysis on the histogram, ViT structure and global loss functions of our proposed approach in Tab.~\ref{tab:ablation_quant}. Kindly note that the structure loss helps in maintaining the structural consistency between the source and protected image while histogram and global losses ensure faithful makeup transfer between reference image and the protected sample at the local and global levels respectively. As expected, removing the global loss ($\mathcal{L}_{\text{glob}}$) increases the FID (Tab.~\ref{tab:ablation_quant}), verifying its importance in preserving the naturalness of the protected sample.

\noindent \textbf{Robust Correspondence Module}:  The robust correspondence module (RCM) is crucial for ensuring a faithful makeup transfer between corresponding regions of the source and reference makeup images. As demonstrated in the qualitative results presented in Fig.~\ref{fig:robust_module}, the absence of the correspondence module leads to makeup artifacts stemming from adversarial toxicity. These artifacts can manifest as misplaced makeup elements or unnatural blending, compromising the overall realism of the protected image. By incorporating the RCM, our approach achieves a more precise and natural-looking makeup transfer, effectively maintaining the identity of the source image while applying the desired makeup style.

\noindent \textbf{Robustness against makeup styles:} We also assess the influence of different reference makeup images on the PSR of the resulting output image. We use five reference makeup images to protect 500 CelebA-HQ images with our method. The results in Tab. \ref{table:robust} indicate a slight variation in the PSR (reflected in the low standard deviation) across different makeup reference images, suggesting that \texttt{DFPP} is robust to a range of makeup styles.

\begin{table}[t]
\centering
\begin{minipage}{.48\linewidth}
  \caption{Quantitative ablative analysis on the histogram, ViT structure, and global losses.}
  \label{tab:ablation_quant}
  \setlength{\tabcolsep}{3pt}
  \centering
  \scalebox{0.85}{%
  \begin{tabular}{l | c|c|c|c }
    \toprule[0.15em]
    \rowcolor{mygray} Metrics & w/o $\mathcal{L}_{\text{hist}}$ & w/o $\mathcal{L}_{\text{ViT}}$ & w/o $\mathcal{L}_{\text{glob}}$ & Overall\\
    \midrule[0.15em]
    FID $\downarrow$ & 30.64  & 30.91 & 31.07 & 29.81 \\
    PSR $\uparrow$ & 69.31 & 70.22 & 69.24 & 69.34 \\
    \bottomrule[0.1em]
  \end{tabular}}
\end{minipage}%
\hspace{0.04\linewidth}%
\begin{minipage}{.48\linewidth}
  \caption{\small Average PSR of \texttt{DFPP} on CelebA-HQ images with 5 reference makeup images provided by \cite{hu2022protecting}. Std. denotes standard deviation.}
  \label{table:robust}
  \setlength{\tabcolsep}{4pt}
  \centering
  \scalebox{0.8}{
  \begin{tabular}{l | c  | c  | c |c |c|c}
    \toprule[0.15em]
    \rowcolor{mygray} & Ref-1 &Ref-2& Ref-3& Ref-4 & Ref-5 & Std. $\downarrow$  \\
    \midrule[0.15em]
    PSR  &  62.2 & 60.4  & 63.6 & 66.0 & 60.8& 2.03 \\
    \bottomrule[0.1em]
  \end{tabular}
  }
\end{minipage}
\end{table}

\begin{wrapfigure}{r}{0.6\textwidth}
\vspace{-0em}
\begin{minipage}{\linewidth}
  \centering
  \includegraphics[width=\linewidth]{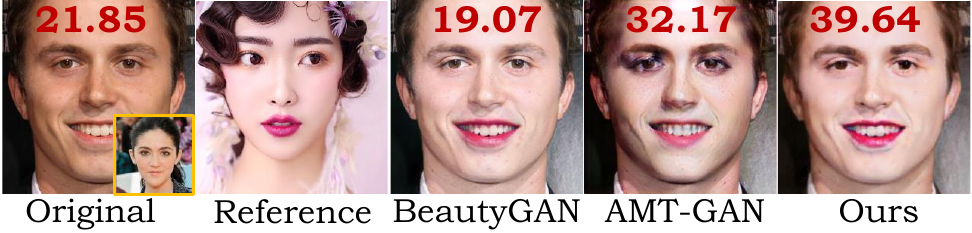}
  \caption{Qualitative comparison with BeautyGAN. The red text represents confidence score (higher is better) by a commercial API, when matching the protected image with the false target identity (shown at the bottom of the original image). Note that although BeautyGAN faithfully transfers makeup, it does not offer facial privacy protection.}
  \label{fig:title_fig}

  \vspace{0em} 

  \captionof{table}{\small PSR for multi-targets and w/o ensemble (ens.) (Rank-1-Targeted).}
  \label{tab:tipim}
  \setlength{\tabcolsep}{3pt}
  \scalebox{0.6}{%
  \begin{tabular}{l | c|c|c|c }
    \toprule[0.15em]
    \rowcolor{mygray} Method & ens. 1-target & ens. 4-targets & ens. 10 targets & w/o ens. (10 targets) \\
    \midrule[0.15em]
    TIP-IM & 8.0  & 23.4  & 69.4 & 55.4 \\
    \rowcolor{cyan!20} Ours & 10.2 & 26.6 & 71.8 & 58.4 \\
    \bottomrule[0.1em]
  \end{tabular}}
\end{minipage}
\vspace{-2em}
\end{wrapfigure}

\noindent {\textbf{On the Role of Makeup:}} We further present a visual quality comparison between facial privacy protection methods (\texttt{DFPP} and AMT-GAN) and solely makeup transfer methods like BeautyGAN~\cite{li2018beautygan}. As depicted in Fig.~\ref{fig:title_fig}, BeautyGAN proficiently transfers makeup but falls short in providing protection. In contrast, our method maintains image quality on par with BeautyGAN while achieving a PSR higher than AMT-GAN. This demonstrates \texttt{DFPP}'s ability to strike a crucial balance between aesthetic makeup application and effective privacy protection. The high-quality result highlight the potential of our approach in real-world scenarios where both visual appeal and privacy safeguards are essential.

\noindent {\textbf{Multi-target setting}}: Our method adopts the single target and ensemble settings from the recent makeup image-based facial privacy approach~\cite{hu2022protecting}. In our main experiments, we showed results with this setting. For TIP-IM, we relied on its official implementation to run the experiments in the AMT-GAN settings (4 FR models, single target, ensemble). Here, we also provide the results in TIP-IM settings that is with multi-target and without ensemble. As depicted in Tab.~\ref{tab:tipim}, our \texttt{DFPP} approach consistently outperforms TIP-IM in both multi-target and non-ensemble configurations.  This demonstrates the versatility and robustness of \texttt{DFPP} across different operational settings. We deploy IRSE50 as a black-box model, with MobileFace, IR152, and FaceNet acting as surrogates in the ensemble (ens.) setting, and only MobileFace as a surrogate in w/o ensemble setting.

\vspace{-1em}
\section{Conclusion}

To conclude, we propose a test-time optimization approach for facial privacy protection through adversarial makeup transfer. We leverage the structure of the randomly-initialized decoder network as a strong prior, which is fine-tuned through structural and makeup consistency losses to produce a protected sample. The output protected sample is perceptually similar to the source image but contains adversarial makeup of the reference image to evade FR models. Compared to existing baseline methods, our approach offers enhanced naturalness and a superior protection success rate. This adaptability, combined with its ability to function without relying on high-quality pre-trained generative models trained on massive face datasets, makes our approach a promising solution for real-world privacy challenges posed by facial recognition technologies.

\clearpage  

%
%

\bibliographystyle{splncs04}
\bibliography{main}

\clearpage
\noindent \textbf{\Large Supplementary: Makeup-Guided Facial Privacy Protection via Untrained Neural Network Priors} 

\section{Datasets Description and Preprocessing} \label{sec:datasets}

This section outlines the datasets and processing steps used in our experiments:
\begin{itemize}
    \item \textbf{CelebA-HQ}\cite{karras2017progressive}: High-resolution dataset (1024 × 1024) with 30,000 images. We use 1000 images of different identities as provided by Hu \textit{et al.}~\cite{hu2022protecting}.

    \item \textbf{LADN}\cite{gu2019ladn}: Makeup-based dataset used for impersonation attacks in face verification. We use 332 non-makeup images split into four groups, each targeting one of four identities provided by Hu \textit{et al.}~\cite{hu2022protecting}.

    \item \textbf{LFW}~\cite{huang2008labeled}: Face identification dataset with 13,233 images and 5,749 identities. Used for face verification (dodging) and face identification (impersonation and dodging). We select 500 pairs, each of the same identity.
\end{itemize}

\noindent We use CelebA-HQ and LADN for impersonation attacks in face verification, and CelebA-HQ and LFW for other settings. This combination demonstrates our method's generalization across high-quality (CelebA-HQ) and low-quality (LFW) images.

\noindent \textbf{Preprocessing}: We use MTCNN~\cite{zhang2016joint} for face detection, cropping, and alignment before input to FR models. Additional preprocessing follows Tov \textit{et al.}~\cite{tov2021designing} for latent code initialization.

\section{Dodging Attack under Face Verification} \label{sec:dodge}

Tab.~\ref{table:verification_dodging} presents PSR results for dodging attacks in face verification under a black-box setting. We use 500 randomly selected subjects, each with a pair of faces. We compare \texttt{DFPP} only with the state-of-the-art noise-based method TIP-IM, as Adv-Makeup\cite{DBLP:conf/ijcai/YinWYGKDLL21} and AMT-GAN~\cite{hu2022protecting} are designed specifically for impersonation attacks. Our method \texttt{DFPP} demonstrates superior performance, achieving an absolute gain of over $5\%$ compared to TIP-IM.

\begin{table}
\centering
\caption{\small Protection success rate (PSR \%) of \textit{black-box} dodging attack under the face verification task. For each column, the other three FR systems are used as surrogates to generate the protected faces. We exclude Adv-Makeup~\cite{DBLP:conf/ijcai/YinWYGKDLL21} and AMT-GAN~\cite{hu2022protecting} as they are trained specifically for impersonation attacks.}
\label{table:verification_dodging}
\vspace{-2mm}
\setlength{\tabcolsep}{2.0pt}
\scalebox{0.8}{
\begin{tabular}{l || c c c c || c c c c || c }
\toprule[0.15em]
\rowcolor{mygray} \textbf{Method} & \multicolumn{4}{c||}{\textbf{CelebA-HQ}}&\multicolumn{4}{c||}{\textbf{LFW}}&\multicolumn{1}{c}{\textbf{Average}} \\
\rowcolor{mygray}  & IRSE50 & IR152 & FaceNet& MobileFace & IRSE50 & IR152 & FaceNet& MobileFace &  \\
\midrule[0.15em]
$\text{TIP-IM}$~\cite{yang2021towards}  & 71.2 & 69.4 & 88.2& 59.0 & 71.8 & 76.1 & 80.6 & 62.9 & 72.4 \\
\midrule
\rowcolor{cyan!20} \texttt{DFPP} (Ours) & \textbf{79.0} &\textbf{80.4} & \textbf{91.7} & \textbf{62.1} & \textbf{76.2} & \textbf{78.6} &\textbf{85.3}  & \textbf{70.9} & \textbf{78.0}  \\
\bottomrule[0.1em]
\end{tabular}}
\end{table}

\section{Results for Face Identification} \label{sec:identification}

The PSR on CelebA-HQ dataset for dodging and impersonation attacks are provided in Tab.~\ref{table:identificationceleba}. For this experiment, we randomly select 500 subjects, each having a pair of faces. We assign one image from each pair to the gallery set, and the other to the probe set. Both impersonation and dodging attacks are conducted on the probe set. We excluded AMT-GAN and Adv-Makeup from both tables, as they are specifically trained for the face verification task.

\begin{table}[t]
\begin{center}
\caption{\small Protection success rate (PSR \%) of \textit{black-box} dodging (top) and impersonation (bottom) attacks under the face identification task for CelebA-HQ dataset~\cite{huang2008labeled}. For each column, the other three FR systems are used as surrogates to generate the protected faces. R1-U: Rank-1-Untargeted, R5-U: Rank-5-Untargeted,  R1-T: Rank-1-Targeted, R5-T: Rank-5-Targeted. }
\label{table:identificationceleba}
\vspace{-2mm}
\setlength{\tabcolsep}{6pt}
\scalebox{0.8}{
\begin{tabular}{l || c c || c c || c c || c c || c c }
\toprule[0.15em]
\rowcolor{mygray} \textbf{Method} & \multicolumn{2}{c||}{\textbf{IRSE50}}&\multicolumn{2}{c||}{\textbf{IR152}}&\multicolumn{2}{c||}{\textbf{FaceNet}}&\multicolumn{2}{c||}{\textbf{MobileFace}}&\multicolumn{2}{c}{\textbf{Average}} \\
\rowcolor{mygray}  & R1-U & R5-U & R1-U & R5-U & R1-U & R5-U & R1-U & R5-U & R1-U & R5-U  \\
\midrule[0.15em]
$\text{TIP-IM}$~\cite{yang2021towards} & 79.6 & 61.2& 62.9 & 42.8 & 46.2 & 27.8 & 81.9 & 76.7 &67.6 & 52.1\\
\midrule
\rowcolor{cyan!20} Ours & \textbf{85.2} &\textbf{68.8} & \textbf{67.4} & \textbf{45.3} & \textbf{54.1} & \textbf{29.5} & \textbf{91.2} & \textbf{81.0} & \textbf{74.5} & \textbf{56.1}  \\
\midrule[0.15em]
\rowcolor{mygray}  & R1-T & R5-T & R1-T & R5-T & R1-T & R5-T & R1-T & R5-T & R1-T & R5-T  \\
\midrule[0.15em]
$\text{TIP-IM}$~\cite{yang2021towards} &16.2  & 51.4& 21.2 &56.0 & 8.1 & 35.8 & 9.6 & 24.0 &13.8 & 41.8\\
\midrule
\rowcolor{cyan!20} Ours & \textbf{22.1} &\textbf{60.2} &\textbf{23.7} & \textbf{62.4} & \textbf{11.8} &  \textbf{37.4} & \textbf{11.1} &\textbf{27.6} &\textbf{17.2} & \textbf{46.9} \\
\bottomrule[0.1em]
\end{tabular}}
\end{center}\vspace{-1.5em}
\end{table}

\section{Results for Tencent Yunshent API}

\begin{figure}[] 
\centering
\includegraphics[clip, trim=0cm 0.1cm 0cm .1cm,width=0.5\textwidth]{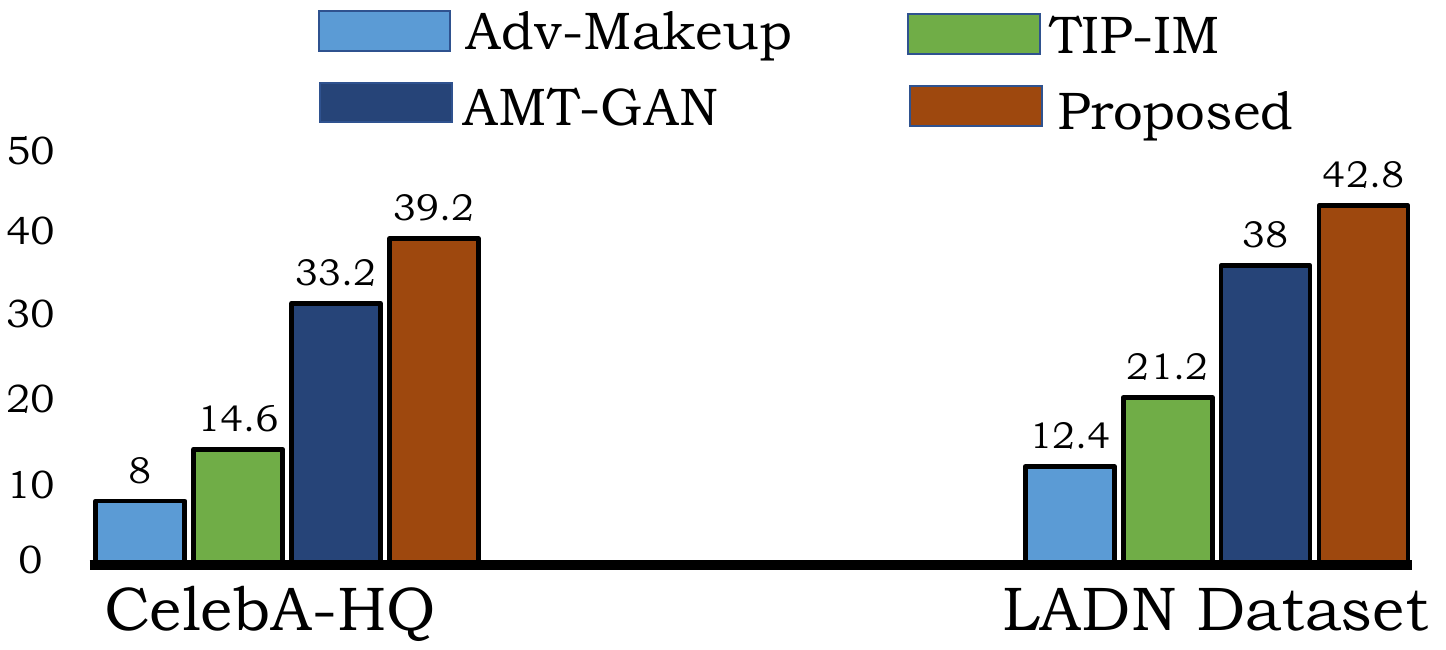}
\vspace{-1em}
\caption{\small Average confidence scores (where higher scores are preferable) from commercial API Tencent Yunshentu  for impersonation attacks within the face verification framework. The \texttt{DFPP} method consistently outperforms these approaches.}
\label{fig:ten}
\end{figure}

We evaluate \texttt{DFPP}'s effectiveness against the commercial API Tencent Yunshentu, operating in verification mode for impersonation. This API returns confidence scores from 0 to 100, with higher scores indicating greater similarity between two images. As the training data and model parameters of this proprietary FR system are undisclosed, this test effectively simulates a real-world scenario. We protect 100 faces from the CelebA-HQ dataset using both baseline methods and \texttt{DFPP}. Fig. \ref{fig:ten} illustrates the average confidence scores returned by Tencent Yunshentu for these protected images. The results clearly demonstrate \texttt{DFPP}'s superior PSR compared to the baselines, underscoring its effectiveness in real-world applications.

\section{Reference Makeup Images} \label{sec:experiments}

\begin{figure}[t]
    \centering
    \includegraphics[width=0.15\textwidth]{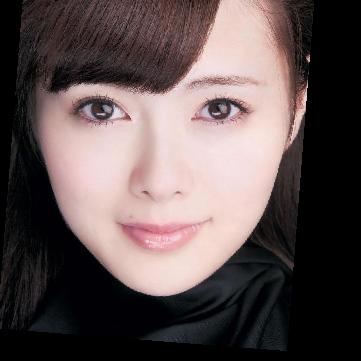}
    \includegraphics[width=0.15\textwidth]{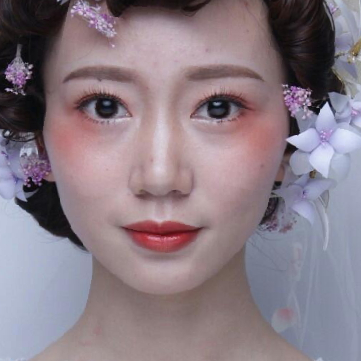}
    \includegraphics[width=0.15\textwidth]{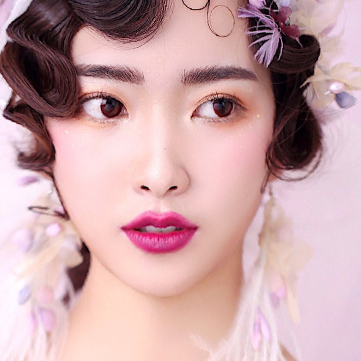}
    \includegraphics[width=0.15\textwidth]{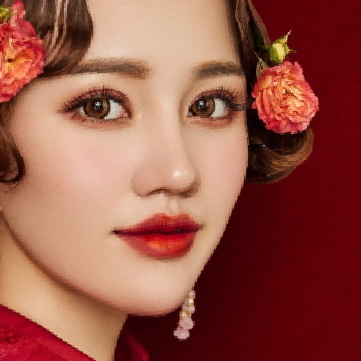}
    \includegraphics[width=0.15\textwidth]{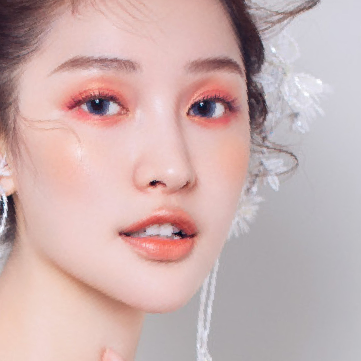} \\
    \includegraphics[width=0.15\textwidth]{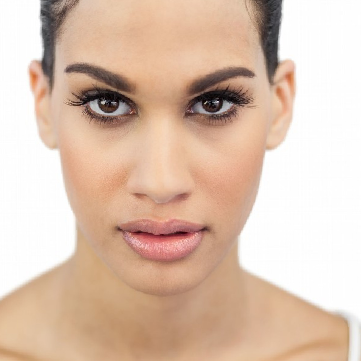}
    \includegraphics[width=0.15\textwidth]{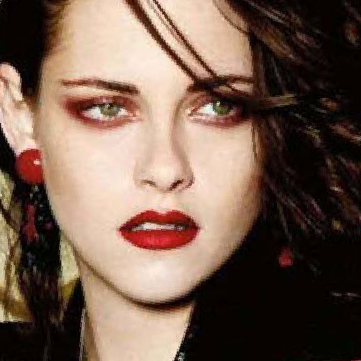}
    \includegraphics[width=0.15\textwidth]{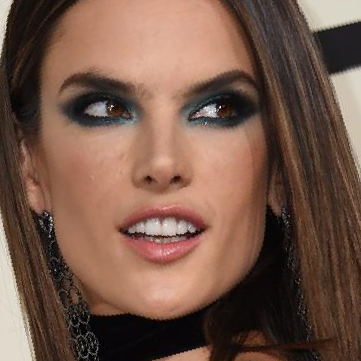}
    \includegraphics[width=0.15\textwidth]{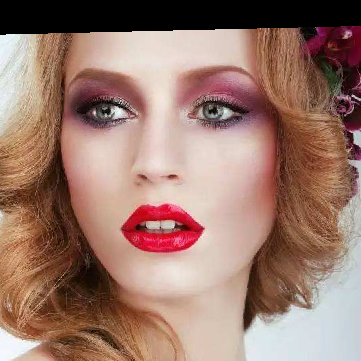}
    \includegraphics[width=0.15\textwidth]{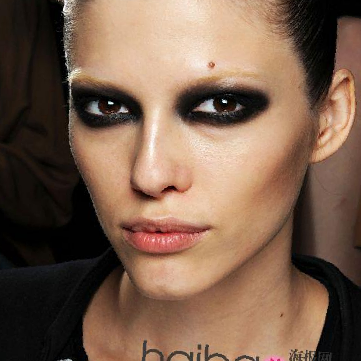}
    \caption{Reference makeup images used by \cite{hu2022protecting} to adversarially transfer makeup to the source image.}
    \label{fig:makeup}
\end{figure}

In all our experiments, we utilize the reference makeup images shown in Fig.~\ref{fig:makeup}. These diverse images, provided by \cite{hu2022protecting}, represent a wide range of makeup styles, from subtle to dramatic. Our results are averaged over these ten reference images, ensuring a comprehensive evaluation of our method's performance across various makeup styles.

\end{document}